\documentclass[a4paper, 11 pt]{article} 
\usepackage[utf8]{inputenc}
\usepackage{lipsum}
\usepackage{graphicx}
\usepackage{subfigure}
\usepackage{amsmath}
\graphicspath{{images/}}
\usepackage{algorithm,algorithmic}

\usepackage{xcolor}

\title{Controlling Robots using Artificial Intelligence and a Consortium Blockchain}

\author{Vasco Lopes, Lu\'{i}s A. Alexandre and Nuno Pereira
\thanks{This work was partially supported by the Tezos foundation throught a grant for project Robotchain.}
\thanks{The authors are with Departamento de Informática, Universidade of Beira Interior and Instituto de Telecomunicações, Covilhã, Portugal
        {\tt\small \{vasco.lopes, luis.alexandre, nuno.pereira\}@ubi.pt}}%
}
\date{}
\begin{document}

\maketitle
\thispagestyle{empty}
\pagestyle{empty}

\begin{abstract}
Blockchain is a disruptive technology that is normally used within financial applications, however it can be very beneficial also in certain robotic contexts, such as when an immutable register of events is required. Among the several properties of Blockchain that can be useful within robotic environments, we find not just immutability but also decentralization of the data, irreversibility, accessibility and non-repudiation. In this paper, we propose an architecture that uses blockchain as a ledger and smart-contract technology for robotic control by using external parties, Oracles, to process data. We show how to register events in a secure way, how it is possible to use smart-contracts to control robots and how to interface with external Artificial Intelligence algorithms for image analysis. The proposed architecture is modular and can be used in multiple contexts such as in manufacturing, network control, robot control, and others, since it is easy to integrate, adapt, maintain and extend to new domains.

\end{abstract}

\section{INTRODUCTION}
Robots are getting more common, especially in tasks that traditionally were performed by humans. One of those tasks is to aid in autism therapy \cite{Rudoviceaao6760}, where robots assist in the therapy and acquire vital information that leads to the possibility of evaluating the engagement of children undergoing autism therapy \cite{8594177}. 
This is part of Human-Robot Interaction (HRI), where most robots need to be supervised or teleoperated \cite{doi:10.1177/0018720816644364}.This control can take many forms, but it will essentially pass by teleoperation or by supervision of the robot's actions. Another application is the use of robots for surveillance and monitoring of physical spaces \cite{6161683,Lawson2017}. Applications using collaborative robots have also been under increasing development, mainly for work collaboration \cite{doi:10.1177/1729881417716010, bauer2008human}, to monitor workspaces \cite{8470596}, to avoid collisions by using vision to do reactive planning \cite{Dumonteil2015ReactivePO}, or external sensors that use distance from robots to humans to stop the robot \cite{6840201}. Other examples use sensors based on pressure to stop or slow down the robot upon space violation \cite{8118009}. Swarm robotics have also shown to have the potential to disrupt many applications, but networks of robots suffer from the problem of the usage of global information versus local information and control mechanisms \cite{khaldi2015overview}.

Security among robot networks, coordination and control of those networks and assurance that the robots actuate within certain ethical and moral limits are some of the challenges that robotics have to surpass \cite{Yangeaar7650}. Blockchain can be the solution to these problems if used properly. Blockchain is mainly used for cryptographic currency transactions, but it is a very powerful tool, allowing for a decentralization of the data and homogeneous registries among all the peers. Since the first idealization of a method for blockchain by Leslie Lamport in 1998 \cite{Lamport:1998:PP:279227.279229}, to the implementation of this technology to serve as the base for Bitcoin \cite{Nakamoto2008}, it has been used to serve many purposes. Incipient work that integrates this technology with AI has been proposed \cite{blockchain:IA:8, DBLP:journals/corr/abs-1810-00329}. The integration of AI with blockchain has been focused on the development of marketplaces and the use of the data contained in it to do predictive analysis \cite{salah2019blockchain}. Despite the fact that integration of blockchain and robotics is still in its infancy, preliminary work \cite{Ferrer2016} presents the possible benefits of combining these two technologies, especially with swarm robotics and robotic hardware. The predominant benefits presented are the possible use of global information within robotic swarms in a secure and validated way and a faster way to change the behaviour of the network, which will ultimately lead to higher productivity and easier maintenance. In \cite{Ferrer2018}, the authors present a conceptualisation of a possible integration of robotics and blockchain, which consists of a method to share critical data among robots in a secure way. This is presented as a framework to tackle privacy issues regarding using personal data by robots during a Human-Robot Interaction. In \cite{basegio2017decentralised}, an architecture to allocate tasks within multi-agent systems using a private blockchain is proposed. This is built on the idea of having communication and coordination throughout the whole system. This proposal uses blockchain as a ledger and has a platform built outside of it to allocate tasks to the robots. \cite{Teslya18} use knowledge processors and information stored on blockchain to create coalitions of robots. In \cite{strobel2018blockchain}, it is shown how a blockchain with a reputation system can be used to achieve consensus in robotic networks and at the same time, be robust enough to deal with byzantine robots \cite{Strobel:2018:MBR:3237383.3237464}. 

With these conceptualisations and proposals, it is possible to see the power that the integration of blockchain with robotics has. Even though there isn't much work done, it's a growing field and once standards are defined industrial and research applications will grow exponentially.

In the literature, there are many different ways of controlling robots, depending on the information required and the behaviour to be achieved \cite{736776,6334723, 7139020}, however, to change the behaviour of the pre-defined control it is usually required to do the entire control logic again. In this paper, we propose a method that allows the integration of different modules with ease and ensures that the data is securely stored. We demonstrate this proposal by developing a novel way of controlling robots. Our approach uses blockchain as a decentralized ledger, that stores information about robots and coupled sensors, like cameras, and smart-contracts to define the logic of control. By having Oracles processing the data and inserting analytics into the blockchain, specialized smart-contracts can use that information to control a robot. We demonstrate this architecture by controlling a UR3 arm \cite{UR3} in a scenario that replicates a factory pick-and-place task, where the robot needs to pick raw material and place it in a new location. As the proposed model is modular, it is easy to integrate new modules that perform other tasks or different behaviour for the same one, and the method is free of restrictions, which means that it can be used in any robot without requiring the robot to have a specific structure, complexity or price tag or computing power.

The contributions of this work can be summarized as follows:
\begin{itemize}
    \item We provide a short overview of the work being done that integrates blockchain with robotics and possible solutions and challenges.
    \item We provide a brief explanation of how blockchain works, how it is possible to use it with robotic systems and how smart-contracts can be designed for it. 
    \item We demonstrate that it is possible to control robots with smart-contracts, by proposing a generalist approach that can be easily used to control any robot.
\end{itemize}

The  rest  of  the  paper  is  organized  as  follows.  First, we  provide  a  background  on the blockchain technology and explain RobotChain \cite{RobotChain}. Then, we explain the proposed method that enhances the capabilities of robotics over blockchain. Finally, we provide a conclusion and a discussion about the proposed method.

\section{Blockchain and Robotics}

\subsection{Blockchain}

Blockchain is a recent technology that has been disruptive in many industrial and academic fields. The core idea behind the blockchain is to have a model capable of reaching a consensus over a network of computers in which the computers may be unreliable. Blockchain is a digital ledger that is implemented in a decentralized way that allows it to store sequences of blocks. These blocks hold information, which is, in most cases, representations of transactions. This technology was popularized mainly by using it as basis for cryptocurrencies, like Ethereum \cite{wood2014ethereum}. However, it is being used in a panoply of applications thanks to the smart-contract technology implemented over it and because it allows to decentralize data in a secure way. The vast majority of blockchain implementations either use Proof-of-Work (PoW) or Proof-of-Stake (PoS) as a consensus algorithm. PoW requires miners to solve a cryptographic puzzle in order to "mine" a block. In PoS, miners are selected to "mine" a block by different algorithms that take into account the miners stake (number of tokens they hold). There have been many consensus algorithms proposed both in the literature and in the industry that try to tackle problems regarding energy and time consumption on validating blocks. In short, Blockchain properties are: replication, decentralization of the data, irreversibility, accessibility, non-repudiation, time-stamping of transactions and (pseudo) anonymity. These properties ensure that, as the time passes, a block added to the blockchain gets more secure, due to the linked chain of hashes. Every transaction sent to the network is mined and if accepted into the blockchain, it becomes valid and secure.

\subsection{RobotChain}

Our approach uses RobotChain as blockchain, which was initially proposed in \cite{RobotChain}. RobotChain is a system to register robotic events in a trust and secure way and it's based on the blockchain developed by Tezos \cite{Goodman2014,Goodman2016} as it incorporates characteristics that are important for systems that are intended to work in sensitive environments. The first characteristic is the support of easier formal verification of smart-contracts, which are written in Michelson, a stack-based language, or in Liquidity and then compiled to the latter. This is very important when working with robots, as it provides security that the code will do what it was defined to do. The second characteristic is the self-amending property that allows changes to be performed on the blockchain by voting on-chain, without the need to conduct hard-forks when a core feature needs to be changed. Finally, the last characteristic is the consensus algorithm, Delegated Proof-of-Work, where the energy and time needed to validate transactions are more efficient than PoW, which is critical when dealing with huge amounts of data. RobotChain is a consortium blockchain \cite{8234700}, which consists of a private blockchain where it's unknown if the nodes can be trusted, which is useful when dealing with factories with multiple robots from different manufacturers. For this type of blockchain, the self-amendment process is useful when there is a need to change the protocol without the need for major changes. The need for registering robotic events in such a way is due to the fact that a human could alter logs, trying to blame an innocent robot for actions that did not performed. This was one of the reasons behind the development of RobotChain. It was created by cloning the main net of Tezos and performing changes to adequate it to our requirements. A transaction parameter named "Transaction Description" was added, providing a field to document the transaction or a smart contract call, allowing texts on transactions. The limits that smart-contracts have were also changed, to allow them to storage data and finally, the genesis keys were changed to ensure security. As RobotChain is a modular approach, applications can be built on top of it, either by inserting smart-contracts with different logics and actions onto the blockchain or by using Oracles. 

\section{Proposed Method}

\begin{figure*}
  \centering
  \includegraphics[scale=0.73]{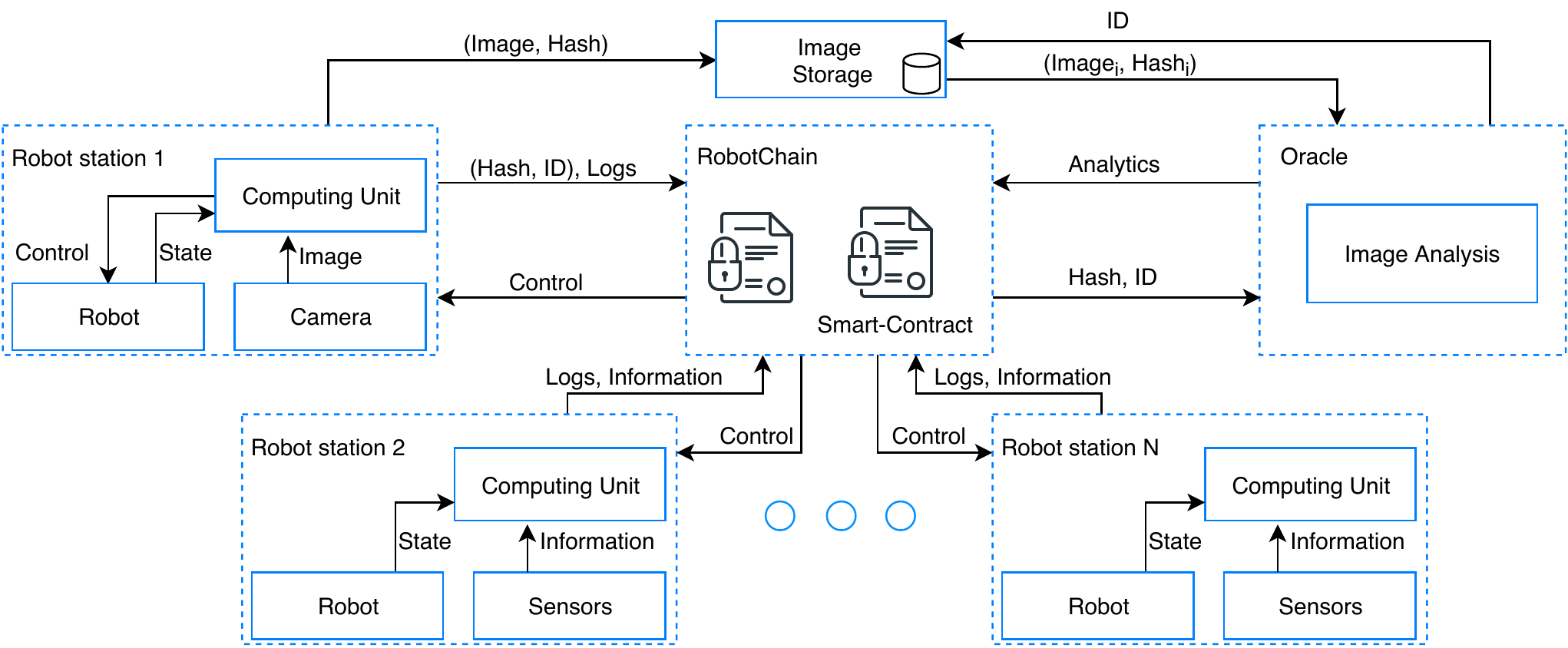}
  \caption{Architecture of the proposed method.}
  \label{fig:architecture}
\end{figure*}

\subsection{System Description}

\begin{figure}
  \centering
  \includegraphics[scale=0.25]{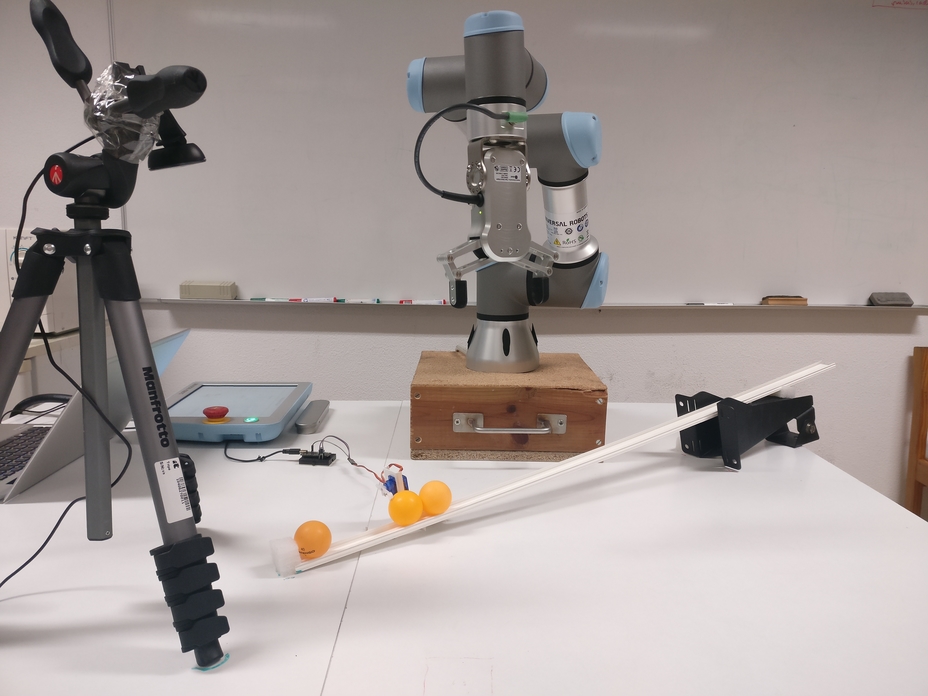}
  \caption{Scenario that represents a factory environment pick and place task. In this, the UR3 arm needs to pick objects from the end of the track-line and place them at the beginning.}
  \label{fig:scenario}
\end{figure}

\begin{figure*}[!]
\centering
\includegraphics[height=2.65cm, width=2.85cm]{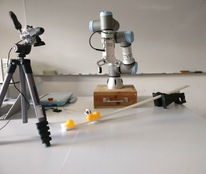}
\includegraphics[height=2.65cm, width=2.85cm]{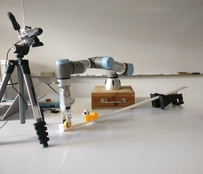}
\includegraphics[height=2.65cm, width=2.85cm]{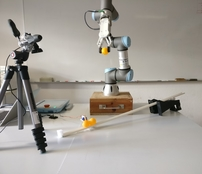}
\includegraphics[height=2.65cm, width=2.85cm]{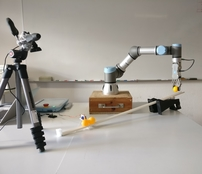}
\includegraphics[height=2.65cm, width=2.85cm]{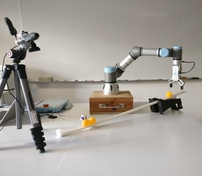}
\includegraphics[height=2.65cm, width=2.85cm]{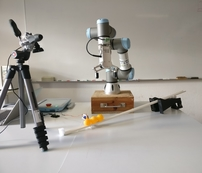} 


\caption{Pick and place task. UR3 arm picks and places a ping-pong on the track and then goes to the home position because the is nothing more to pick.\label{fig:eventsrobot}}
\end{figure*}

We propose a method to control robots in a secure way, which is verified and can't be changed without permission. For this, we use RobotChain as a decentralized ledger, capable of storing robotic events in a secure and fast way. This is useful for having global information about a network and to register information about robots, which contains possible errors that the robots have or perform. We store the information in a smart-contract storage and we control the robots with smart-contracts as well. This is a generalist approach that can be used with any robot as long as the robot allow external commands to perform changes on its' behaviour, e.g., changes on the speed. The smart-contract is useful because it contains all the logic for the changes to be made and it is not controlled by any party, in the sense that once published into the blockchain, it is impossible to change the code contained in it and it triggers events without the need of human actions. Another important property is the capability for having different smart-contracts with different control logics, meaning that for different environments or tasks, we can allocate one or more smart-contracts to different types of control of the robot or to other aspects that are found important, e.g., communication with other parties. We demonstrate this approach by creating a scenario that simulates a factory environment in which a robotic arm needs to pick a raw material from one conveyor and place it on another one. In the created scenario, the robotic arm is a UR3, the raw material are orange ping-pong balls and the robot needs to pick it from the end of a track-line and place it at the beginning. We have a small motor that blocks the balls from reaching the end of the track for $n$ seconds to represent different arrivals at the robot workspace. This is useful to test how the system performs under different raw material arrival intervals. Denote that the motor that blocks the balls is controlled by an Arduino and it only lets one ball pass each $n$ seconds. This scenario can be seen in Fig. \ref{fig:scenario} in which there is one ball on the pick zone and two balls "waiting". The complete sequence of movements in this task can be seen in Fig. \ref{fig:eventsrobot}, where the robot picks one ball, places it in the beginning of the track and stops the movement because it is informed to do so since there are no more balls to be picked. The final image represents the home position to which the robot returns if there is nothing more to pick.

The architecture designed for the proposed method is shown in Fig. \ref{fig:architecture}. This consists of RobotChain as a ledger that contains smart-contracts that execute code to define a robot state. The information contained in the RobotChain is sent by a controlling unit that receives information from the UR3 arm state, which includes position, velocity and effort and an image from a USB camera, placed on a tripod, looking to the place where the balls need to get picked. In our scenario, both connected to the same computing unit, but this does not influence the processing or the blockchain, as they are treated separately. As the blockchain can rapidly increase in size \cite{8590956}, we only store on the blockchain, the robotic events (logs) and a tuple comprised of a Hash and an ID that represents the image. The image itself is stored in a database. As the information stored in the blockchain can't be changed, the Hash of the images adds security to the database, which can be easily corrupted if not created properly. The information on the blockchain is then used by a trusted external Oracle, which can be placed in the Cloud or in a central server. In our approach, the Oracle processes the images to detect how many balls are present. This information is then sent to a smart-contract where it is defined if the robot should stop at the home position, which means that there is no material to pick, or either slow down or speed up, meaning that there are few or many materials to pick, respectively. 

\subsection{The Oracle}

\begin{figure}
    \centering
    \subfigure[Gray]
    {
        \includegraphics[width=0.98in]{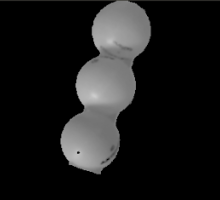}
        \label{fig:first_sub}
    }
    \subfigure[Canny Edge]
    {
        \includegraphics[width=0.98in]{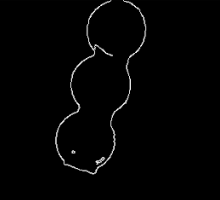}
        \label{fig:second_sub}
    }
    \subfigure[Detection]
    {
        \includegraphics[width=0.98in]{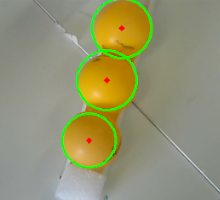}
        \label{fig:third_sub}
    }
    \caption{Different stages of the ball detection algorithm performed by the Oracle.}
    \label{fig:detection}
\end{figure}

The Oracle processes the images by first converting the RBG image to HSV. After this, all the values outside the orange colour space are eliminated. Then, the image is converted to grey by performing a bit-wise operation between a blur of the original image two times with the image from the last step as mask. Then it uses Canny Edge Detector to detect edges and finally, performs Hough Transform to detect circles. This process can be seen in image \ref{fig:detection}, where it is shown the different image stages that lead to the detection of 3 ping-pong balls.

\subsection{The Smart Contract}

\floatname{algorithm}{Procedure}
\begin{algorithm}[H]
\renewcommand{\thealgorithm}{1}
 1: Version Number\\
 2: Definition of data structures (type)\\
 3: Definition of the storage structure\\
 4: Init method\\
 5: Entry points
 
 \caption{Smart-Contract Structure in Liquidity\label{structure:sc}}
\end{algorithm}

Despite the fact that smart-contracts are usually used to do legal contracts or currency transactions, they can be useful to allow the interaction of Oracles with the blockchain, automatically performing actions and to help nodes of a network obtain global information.

In procedure \ref{structure:sc}, we present the typical structure of smart-contracts written in Liquidity. First, it is required to define the version of Liquidity used, for example: \texttt{[\%\% version 0.5]}. Then, it is possible to define data structures by using the 'type' syntax. An example of this is: \texttt{type oracles = key\_hash list}. This step is optional, as one can decide that it's not useful to have data structures defined. The third step is the definition of the storage variable. This step is done similarly to step 2) but it's required that the structure has the name 'storage', for example: \texttt{type storage = oracles}. The fourth step is optional, even though it serves as an initializer for the storage. This method is defined by the \texttt{init} keyword and it is called only one time which is when the smart-contract is deployed. The way this method initializes the storage is by returning the values to initialize the storage, example: \texttt{let\%init init\_oracles (id : key\_hash) = ([id])}. The final requirement in a typical Liquidity structure are the entry points. These points are methods that are used to perform some logic and have actions inside of the smart-contract or on the blockchain. It is possible to have multiple entry points, and when working with Oracles, it is advised to create an entry point for each type of task or user that is allowed to call that smart-contract. Entry points are defined with the \texttt{entry} keyword and return a list of operations and the storage. The signature of an entry point is defined by the \texttt{entry} keyword, the parameters received and the name of the storage variable in that context, for example: \texttt{let\%entry main ((log : string), (oracle : key\_hash)) storage = ( [], storage )}. Note that there are more options that can be used, but in the context of integration between Robotics and blockchain, the presented steps are enough to perform rather complex tasks with smart-contracts.

Our approach uses smart-contracts to register robotic events and to define the logic for robotic control. Specifically to the scenario created, the information of how much material is waiting to be picked up and the robot state, from which it is possible to infer if the robot is transporting anything or not, are used by smart-contracts that define the robot velocity. This velocity is defined in seconds per movement. We defined a simple function on the smart-contract to demonstrate the proposed method. The smart-contract method to define the number of balls to be transported is:

\begin{equation}
    x = ImageAnalysis() + Transporting()
\end{equation}
where $ImageAnalysis()$ is the information that the trusted Oracle (trusted in the sense that only it can insert information onto the smart-contract) inserts into the contract and $Transporting()$ is a function inside the contract that adds 1 if the robot is transporting a ball and 0 otherwise.

The velocity is defined by:

\begin{equation}
    v = \begin{bmatrix}\frac{maxSpeed + meanSpeed}{x}\end{bmatrix}
\end{equation}
where $maxSpeed = 2$ and $meanSpeed = 4$. This enforces that the slowest and the fastest velocities per movement are six and two seconds, respectively, when working with a maximum number of balls to be picked equal to three.


\section{Discussion}


With the architecture proposed we show how it is possible to integrate robotics and blockchain. Even though we showcased a rather simplistic velocity control, it is a novel way of controlling robots that can be adapted to different situations and scenarios and that can also be tuned to ensure that the temperature and other values of the robot don't exceed a threshold in order to avoid failures. This approach is robust to changes due to the fact that no one can alter the control logic without permission, which is imposed by the smart-contract. The blockchain acts as a ledger that securely stores data across a possible untrusted network and, with the capability of interacting with Oracles, external process of the data can be performed, which can integrate many forms, but more importantly, it can bring AI to the blockchain, which is the case on this paper. With our proposal, it is possible to have multiple robots working in different tasks and have a unified system to control them which also presents the benefit of having global information of the whole network that can be used upon request. It is also possible to do other kinds of tasks, as it is a modular approach and can be extended to other tasks and even to multiple Oracles with ease.

The constraints of our approach are: first, how easy it is for the blockchain size grow, that's why we store an image representation instead of the image itself on the blockchain, but this can be improved by removing unnecessary information in transactions. Second, the validation speed, which is the greatest bottleneck of most blockchain approaches. If there are thousands of transactions per second, the time to get an answer to change behaviour can slow down. However, this can be improved by changing the consensus algorithm to one that removes the need for computation in order to validate blocks, and if the environment where the blockchain is deployed is a private one like a factory, the connection speed will help, since it has less communication passing through it.


In table \ref{tab:exp} we present the velocities of the robot at different times from 4 different experiences, where the velocities are in seconds per movement (speed of the robot) and were retrieved at different times (Time column). The velocity values equal to zero mean that the robot is stopped and waiting for material to pick. The experiments started with the same scenario, 3 balls to be picked in the beginning, which means that the initial velocity of the robot is defined by the aforementioned method to 2 seconds per movement. What differs in the experiences is the time, $n$, that the motor takes to open and allow one ball to pass, representing the arrival of a new material to be picked by the robot. The experiences conducted were: A: $n=5s$, B: $n=15s$, C: $n=40s$, D: initial $n=20s$ and incremented by $5s$ every iteration (each time it opens). The values shown are within a 5 minute frame because in all the experiments the values repeat themselves, with exception of experiment D, in which the robot tends to stop more often due to the fact that $n$ keeps on being incremented. The table shows that the proposed method is indeed capable of controlling the robot velocities, which happens in real-time. One property that can be seen in the table, is that due to the fact that the method adjusts the velocities depending on the rhythm of incoming materials, the velocities tend to suffer few changes. For instance, in experience A, even though the velocity stays constant for large periods of time, the method keeps enforcing that the robot should be at a determined velocity, depending on the number of balls to be transported. This is important to keep logs of the robot's and to enable it to adjust to sudden changes.

\begin{table}[]
\centering
\caption{Values of the velocity, in seconds per movement, for the four conducted experiments.}
\label{tab:exp}
\begin{tabular}{ccccc}
\hline
\multicolumn{1}{l}{Time (s)} & A & B & C & D \\ \hline
10                       & 2 & 3 & 3    & 3 \\
20                       & 3 & 3 & 3    & 3 \\
30                       & 3 & 3 & 6    & 3 \\

50                       & 3 & 3 & 6    & 6 \\
75                       & 3 & 3 & 6    & 6 \\
100                       & 3 & 3 & 6    & 6 \\
125                       & 3 & 3 & 6    & 6 \\
\end{tabular}
\begin{tabular}{ccccc}
\hline
\multicolumn{1}{l}{Time (s)} & A & B & C & D \\ \hline

150                       & 3 & 2 & 0    & 0 \\
175                       & 3 & 3 & 6    & 0 \\
200                      & 2 & 3 & 0    & 6 \\
225                      & 2 & 6 & 3 & 0 \\
250                      & 2 & 3 & 6 & 6 \\
275                      & 2 & 6 & 0 & 0 \\
300                      & 2 & 6 & 6 & 0 \\
\end{tabular}
\end{table}

\section{Conclusions}

This paper describes how blockchain can be integrated with robotics by using RobotChain as a decentralized ledger. The proposed architecture is capable of registering robotic events and uses Oracles for processing acquired data (in our
example, images) and controls robots by using smart-contracts. By using such a modular architecture, it is possible to insert new modules that can either serve as Oracles and do any type of data processing, such as image or log analysis, or new smart-contracts to act upon the system in a secure way. This system ensures that no one can change past states that were inserted into the blockchain and that the "output" of the smart-contracts is always free of human changes, meaning that no one can alter a smart-contract logic once it is on the blockchain. Our proposal shows that it is possible to integrate blockchain with robotics and that this integration has advantages beyond having only a way to register and transmit information. Even though we demonstrated the proposed method on a mechanism to control a robot, it can be used in other contexts, such as: distributing tasks to a network of robots; having a mechanism where robots can ask for help in their task if they can't perform it or if they have no information about a specific requirement, for example, identifying humans or objects in images (the robot may not know what are the objects but other robots may); detecting productivity and problems in robots, which can be useful in factories or to monitor other aspects of the stations connected to the blockchain.

The blockchain can also be further tailored, for instance by including voting consensus for swarm robotics agreements and droping the token to further speedup the validation process or replace it for a reputation system which could be useful for task management and consensus, since monetary value is no longer meaningful in consortium blockchains inside private facilities, such as factories.






\bibliographystyle{plain}
\bibliography{bibliografia}

\end{document}